\documentclass[letterpaper]{article}
\usepackage{aaai}
\usepackage{times}
\usepackage{helvet}
\usepackage{courier}

\usepackage{natbib}

\usepackage{graphicx}
\usepackage{multirow}
\usepackage{gensymb}
\usepackage{amsmath}
\usepackage{footmisc}

\usepackage{hyperref}

\usepackage{tabularx,multirow,booktabs,blindtext}
\usepackage{xcolor}
\usepackage{amssymb}
\usepackage{soul}
\usepackage{tabularx}
\usepackage{array}
\usepackage{longtable}
\usepackage{lscape}
\usepackage{subcaption}
\usepackage{scrextend}
\usepackage{enumitem}
\usepackage{cleveref} 
\usepackage{siunitx}

\begin{document}
%
\title{Rxn Hypergraph: a Hypergraph Attention Model for Chemical Reaction Representation}
\author{Mohammadamin Tavakoli $^{*}$,
    Alexander Shmakov $^{*}$,
    Francesco Ceccarelli, 
    Pierre Baldi \\
    \\
School of Information and Computer Sciences,\\
University of California, Irvine\\
mohamadt@uci.edu\\
}

\maketitle

\begin{abstract}
It is fundamental for science and technology to be able to predict chemical reactions and their properties. To achieve such skills, it is important to develop good representations of chemical reactions, or good deep learning architectures that can learn such representations automatically from the data. There is currently no universal and widely adopted 
method for robustly representing chemical reactions. Most existing methods suffer from one or more drawbacks, such as: (1) lacking universality; (2) lacking robustness; (3) lacking interpretability; or (4) requiring excessive manual pre-processing.
Here we exploit graph-based representations of molecular structures to develop and test a hypergraph attention neural network approach to solve at once the reaction representation and property-prediction problems, alleviating the aforementioned drawbacks. We evaluate this hypergraph representation in three experiments using three independent data sets of chemical reactions. In all experiments, the hypergraph-based approach matches or outperforms other representations and their corresponding models of chemical reactions while yielding interpretable multi-level representations.
\end{abstract}

\noindent Over the past few years, artificial intelligence has refashioned organic chemistry. Numerous problems such as chemical reaction prediction, synthesis route planning, drug design, etc., have been benefited from the advancement of deep learning methods \citep{coley2017prediction, fooshee2018deep, schwaller2019molecular, schwaller2020predicting, yang2019analyzing, tavakoli2021quantum}. For instance, accurate reaction yield predictions would help chemists to choose synthesis routes across high-yielding chemical reactions. As another example, estimating reaction rates \textit{via} deep learning methods could circumvent the time and expense required to experimentally measure the reaction rate of reactions. Not to mention that because the true solution-phase reaction rates are bounded by the rate of molecular diffusion, it is impossible to measure the accurate rate experimentally. Lastly, predicting the outcome of chemical reactions using deep learning methods would automate and accelerate the demanding processes of drug design and discovery. All these problems require accurate predictions of chemical reactions' properties using deep learning models, consequently, necessitating optimizing every aspect of such deep learning models. One important facet that can significantly affect training dynamics and model performance is the representation of the input data. 

In an attempt to find a suitable representation of chemical reactions for input into deep learning models, several methods have been proposed but each of them suffers from certain shortcomings. These shortcomings may be summarized with the following properties: 
\begin{enumerate}
\item \textbf{Lack of universality} Several representations are derived based on predefined pattern matching algorithms \citep{schneider2015development, rogers2010extended}. Since there is no learning process involved in deriving these representations, they cannot be automatically adjusted for different predictive tasks.

\item \textbf{Lack of robustness} Atoms and molecules have no inherent ordering within a chemical reaction. Therefore, a robust representation must be invariant to permutations on atoms and molecules within a given reaction. Methods built upon the text representation of chemical reactions are an example of models with the  lack of robustness \citep{schwaller2021mapping, schwaller2021prediction}. Using such methods, one can obtain different outcomes by only permuting atoms and molecules in the text representation of one single reaction.

\item \textbf{Non-interpretability} It is vital for chemists to understand the reasoning behind a reaction-level prediction. For example, capturing the correlation between the presence and absence of certain functional groups or the interaction between specific electrophiles and nucleophiles would provide useful insight for chemists \citep{kadish2021methyl, mood2020methyl}. Thus it is important that a representation can provide means to interpret the final predictions.

\item \textbf{Need for expensive computations} Lastly, some other representations require hand-crafted implementations and processes in order to be used as the input of a predictive model. These hand-crafted processes usually include running pattern and subgraph matching algorithms \citep{kayala2011learning} or performing massive data augmentations \citep{bjerrum2017smiles} which are extremely time-consuming.
\end{enumerate}

In what follows, we review these methods and evaluate them based on the four mentioned shortcomings. Then we propose an augmented graph representation of a chemical reaction called \textit{rxn-hypergraph} which is designed to fix these shortcomings and improve the proposed methods. Finally, through a set of experiments on the classification and plausibility-based ranking of chemical reactions, we empirically show the viability of our \textit{rxn-hypergraph} representation. 

\section{Related Work}
\label{sec:relatedwork}
One commonly used method for numerically representing chemical reactions was introduced in \citep{schneider2015development}. This representation is called \textit{reactionFP} and it is derived from the fingerprints of the molecules involved in the chemical reaction. The \textit{reactionFP} can be described as follows:
\begin{equation}
    \label{eqn:fp_rep}
        w_1(\sum_{P_{i}}^{}FP(P_{i}) - \sum_{R_{i}}^{}FP(R_{i}))+w_2(\sum_{A_{i}}^{}FP(A_{i}))
\end{equation}
Where $A$, $R$, and $P$ represent the molecular agents, reactants, and products respectively. $w_1$ and $w_2$ are two, potentially learnable, parameters that adjust the contribution between the agent molecules and the reactive molecules within the final representation. $FP(.)$ is a function that outputs a traditional fingerprint of a given molecule such as ECFP4 \citep{rogers2010extended}. Since extracting the traditional molecular fingerprints is based on the presence or absence of predefined patterns and involves no learning process, this representation lacks universality and cannot be adjusted for a variety of reaction-level predictive tasks.

However, \textit{reactionFP} was an early robust representation since Equation \ref{eqn:fp_rep} is trivially invariant to the permutation of the molecules involved in the reaction. Additionally, many traditional molecular fingerprints such as ECFP4 \citep{rogers2010extended} or AtomPair \citep{carhart1985atom} are also invariant to atomic order within molecules, presenting a robust representation. Nevertheless, since the molecular fingerprints are obtained using non-invertible hashing mechanisms \citep{glen2006circular}, this method cannot be easily interpreted to discover high-level patterns after learning. Additionally, according to Equation \ref{eqn:fp_rep}, obtaining \textit{reactionFP} requires extra hand-crafted computations including: (1) an accurate atom mapping between reactants and products to identify the role of molecules in a chemical reaction; and (2) extracting the vectorized form of traditional fingerprints.

Graph-based methods have also been used to extract more informative reaction representations. To predict the outcome of chemical reactions, \citet{fooshee2018deep} used a neural network to rank a set of potential mechanistic reactions based on their thermodynamic plausibility. They form two separate count bit vector representations, one for the reactant and one for the product molecules by recursively counting the predefined paths and trees of different sizes rooted at each atom. Then, through a mutual information feature selection stage, they extract the most informative set of these count bits for a given downstream task. Finally, the chemical reaction may be represented as the difference between the count vector of reactants and the count vector of products. Since this method is based on a set of predefined patterns, it lacks universality in the sense that it cannot capture necessary information for different tasks. Although it is not discussed in \citet{fooshee2018deep}, the mutual information feature selection step can potentially provide an additional interpretable view of the final prediction. However, this requires further expensive computations, not to mention that already massive computation required for obtaining the count bit vectors. 

Finally, the most successful reaction representation makes use of SMIRKS \citep{weininger1989smiles} of chemical reactions \citep{schwaller2019molecular, schwaller2021extraction, schwaller2021mapping}. SMIRKS is a well-defined domain-specific language with special characters and grammatical rules for describing chemical transformations in the SMILES strings \citep{weininger1988smiles}. Authors in \citet{schwaller2021mapping} deployed commonly used methods from natural language processing (NLP) to encode the SMIRKS of chemical reactions into a continuous vector. They train bi-directional encoder representation from transformers (BERT) \citep{devlin2018bert} models to obtain task-specific reaction representations. They also trained large sequence to sequence transformer models for masked language model prediction (MLM) on the reaction SMIRKS. These models may also be used to extract pre-trained representations of any given chemical reactions \citep{pesciullesi2020transfer}. These transformer-based methods are highly accurate across multiple reaction level predictive tasks \citep{schwaller2021prediction, schwaller2020predicting}. Such models provide a universal reaction representation that can be used for numerous reaction-level predictive tasks. Additionally, transformer architectures provide a character-level interpretable framework over the SMILES strings. However, one major downside of these NLP models is that their input, SMILES and SMIRKS strings, are not permutation invariant with respect to the order of atoms and molecules. Depending on the canonical SMILES parsing algorithm, atom labeling, and a few other details, a single reaction may be correctly represented with many different SMIRKS. Some permutation invariance may be recovered through massive data augmentation, where all possible representations of the input reactions are generated and randomly selected during training \citep{bjerrum2017smiles}. It has been shown the performance of transformer models is highly dependent on the data augmentation, where they show surprisingly poor performance without the data augmentation \citep{bjerrum2017smiles, schwaller2019molecular}.

\section{Methods}
In this section, we describe the reasons for constructing the chemical reaction hypergraph (\textit{rxn-hypergraph}) and why this hypergraph would yield an abstract and powerful representation of a chemical reaction. Then we explain the process of constructing the (\textit{rxn-hypergraph}) for a given reactions, and finally, we discuss how to train neural networks using this hypergraph representation of chemical reactions.

\subsection{Why Rxn-hypergraph?}

Graph neural networks and their variants have become the preeminent tool for learning patterns and relations from graph-structured data \citep{duvenaud2015convolutional, schlichtkrull2018modeling, kipf2016semi, tavakoli2020continuous, wieder2020compact, de2018molgan}. The main operation of graph neural networks is to recursively update the representation of nodes using a message-passing scheme only between the nodes and their neighbors. Then a read-out function can be applied to the set of nodes' representations to provide an abstract representation of the entire graph.

There are several essential reaction-level properties where predicting them would be highly beneficial to the entire field of chemoinformatics. However, applying a graph neural network to the graph structure of chemical reactions in its most raw form would not lead to a rich representation that captures different properties of the reaction. Particularly, reactions are a more general form of a graph that consists of multiple disconnected graph components (molecules). The absence of a message-passing route between these components would result in node representations that are independent of the nodes and connections within the other graph components that are involved in the reaction. Consequently, applying any form of a read-out function to these independent node representations would result in a non-informative representation of the entire reaction.

On the contrary, self attentional models (e.g. transformers) have been impressively successful in reaction-level property predictions \citep{schwaller2021prediction}. The input to these models is the SMIRKS representation of a chemical reaction in which atoms are represented by alphabetical characters (tokens). Although this form of input representation is not invariant to the permutation, it provides a suitable structure for the transformer architectures. The key reason behind the success of such models can be found in applying multiple layers of self-attention mechanisms to every pair of input tokens. By this means, the representation of each token will be updated by attending to all other tokens including the atoms within other molecules.

Inspired by this crucial factor in representing a reaction, we form a hypergraph structure of a reaction by constructing efficient message passing routes between every pair of atoms. These routes would improve the representation of the reaction by: (1) enabling message-passing schemes between every possible pair of atoms so the atom representations are updated with respect to all other atoms within the reaction, and (2) providing a learnable read-out function (i.e. pooling mechanism) which can attend to different parts in different levels of the reaction which are informative for a specific predictive task. 

\subsection{Constructing Rxn-hypergraph}
A chemical reaction with $N$ reactants and $M$ product molecules is described by two distinct sets of disconnected graph components $R$ and $P$. $R=\{G_{i}^{r}\}_{i=1}^N$ represents the set of reactant molecules, and $P=\{G_{i}^{p}\}_{i=1}^M$ represents the set of product molecules. Molecule $G_i$ with $n$ atoms (regardless of begin a reactant or product molecule) is a graph $G_{i}=(V_{i},E_{i},A,S)$, where $V_i=\{a_j^i\}_{j=1}^n$ is the set of nodes (atoms) and $E_i=\{(a_u^i,a_v^i)\}$ is the set of edges (bonds). The set of possible labels for the vertices in $A$ correspond to atom types (e.g. C, O), and the set of possible labels for the edges in $S$ correspond to edge types (single, double, triple, and aromatic). The idea behind forming the \textit{rxn-hypergraph} of a chemical reaction is to efficiently construct new message passing routes between these disconnected graphs components (molecules) and form one connected hypergraph to represent the entire reaction. 

\begin{figure}[t]
    \centering
    \includegraphics[width=0.99\linewidth]{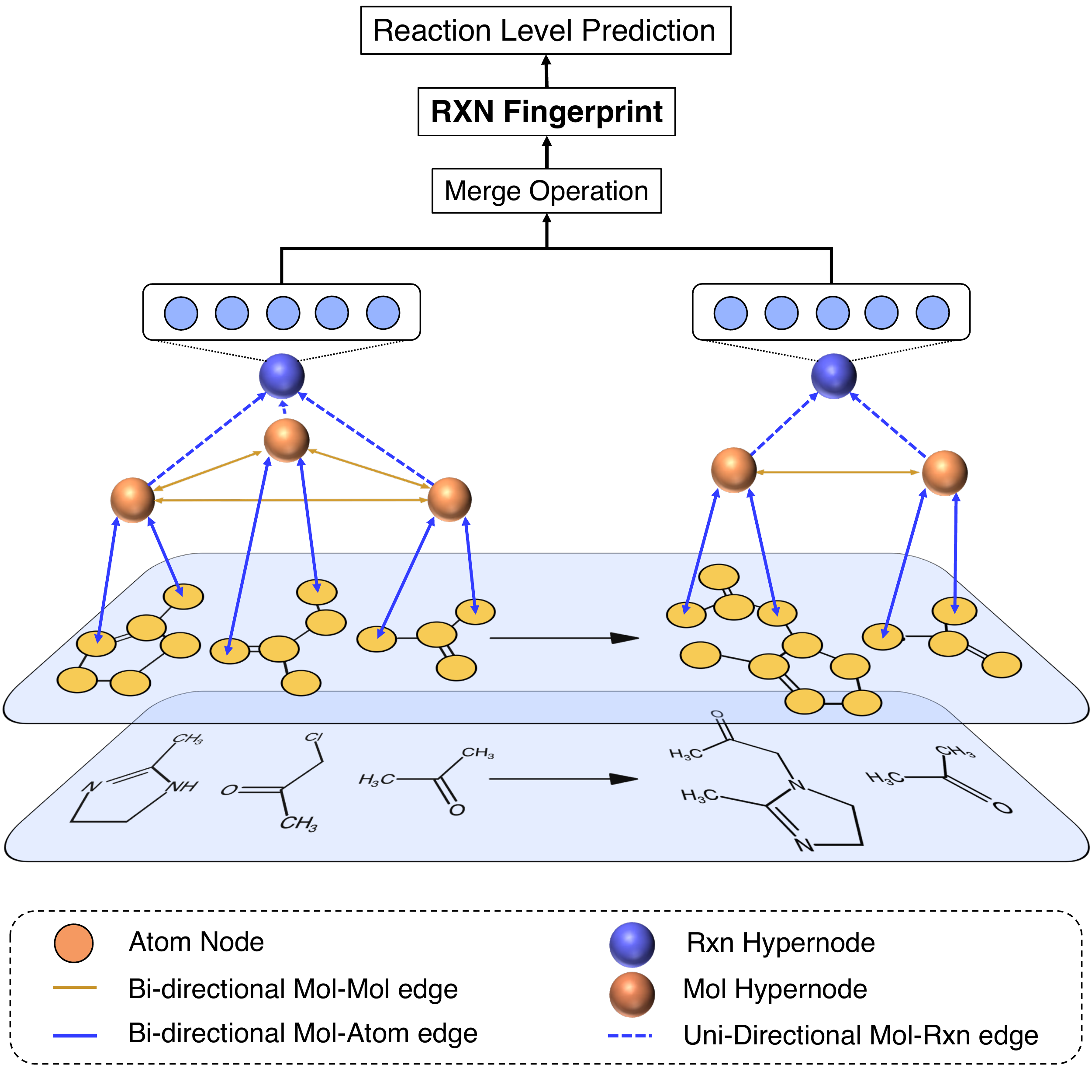}
    \caption{A depiction of the \textit{rxn-hypergraph} corresponding to the reaction at the bottom.}
    \vspace*{-.2in}
    \label{fig:rxn_hypergraph}
\end{figure}

To form this hypergraph, we begin by unifying all the $G_i$s into one graph $G=(V,E,A,S)$ where $V=\bigcup V_{i}^{r}+\bigcup V_{i}^{p}$ and $E=\bigcup E_{i}^{r}+\bigcup E_{i}^{p}$, while $A$ and $S$ remains the same.
For each of the disconnected graph components $G_{i}$ (molecules), we add a hypernode to the graph as a \textit{mol-hypernode} $m_i$. Then we add two types of new edges to the $G$: (1) a set of bidirectional edges connecting every atom to the \textit{mol-hypernode} of their parent molecule, \textit{mol-atom}$=\{(m_i,a_j^i),(a_j^i,m_i)\}$, and (2) a set of bidirectional edges connecting every pair of \textit{mol-hypernodes} on either side of the reaction, \textit{mol-mol}$=\{(m_i,m_j),(m_j,m_i)\}$. The edges of type \textit{mol-mol} would form two fully connected subgraphs between the \textit{mol-hypernodes} on each side of the reaction. We further augment $G$ by adding two more hypernodes as \textit{rxn-hypernodes} $x^{r}$ and $x^{p}$, one for the reactant and one for the product side of the reaction. Then we add a new type of edge to the graph: a set of unidirectional edges from each \textit{mol-hypernode} to the \textit{rxn-hypernode} of the same side of the reaction, \textit{mol-rxn}$=\{(m_i^{r}, x^{r}),(m_j^{p},x^{p})\}$. This augmented version of graph $G$ is what we refer to as the \textit{rxn-hypergraph}. Figure \ref{fig:rxn_hypergraph} shows a schematic drawing of the \textit{rxn-hypergraph}. It can also be represented it as follows:
\begin{equation}
    \label{eqn:rxn-hypergraph}
    \begin{split}
        &\textit{rxn-hypergraph}=(V^{*},E^{*},A^{*},S^{*})\\
        &\text{where:}\\
        &V^{*}=V \cup \{m_i^r\}_{i=1}^{N} \cup \{m_i^p\}_{i=1}^{M} \cup \{x^r,x^p\},\\
        &E^{*}=E \cup \{(m_i,a_j^i),(a_j^i,m_i)\} \cup \{(m_i^r,m_j^r),(m_j^r,m_i^r)\} \\ 
        & \cup \{(m_i^p,m_j^p),(m_j^p,m_i^p)\} \cup \{(m_i^r,x^r),(m_j^p,x^p)\},\\
        &A^{*}=A \cup \{\textit{mol-hypernode},\textit{rxn-hypernode}\},\\
        &S^{*}=S \cup \{\textit{atom-mol},\textit{mol-atom},\textit{mol-mol},\textit{mol-rxn}\}.
    \end{split}
\end{equation}
\subsection{Relational Graph Attention}
Now that the rxn-hypergraph is formed, there is an intermolecular path of length three ($a_u^i \rightarrow m_i \rightarrow a_v^j \rightarrow m_j$) between every pair of atoms. Thus, we may construct robust contextual node representations by applying more than three layers of graph convolution neural networks to the \textit{rxn-hypergraph} since this would ensure the receptive field of every atom would include all other atoms. Also, the presence of the unidirectional paths between \textit{mol-hypernodes} and a \textit{rxn-hypernode}, provides a form pooling mechanism where the \textit{rxn-hypernode} can fully represent one side of the reaction. Assuming that both \textit{rxn-hypernodes} carry an abstract representation of the reactants and products molecules, merging them into one vector would represent the entire chemical reaction.

To obtain this representation, we model the entire \textit{rxn-hypergraph} as a relational graph where the relations are represented by $S^{*}$ in the previous section. We apply and compare two forms of graph neural networks: relational graph convolution (RGCN) \citep{rgcn} and relational graph attention (RGAT) \citep{rgat, rgatv2}. We summarize the layer-wise operations for each of these graph neural networks.

First, a layer-wise operation of the relational graph convolution can be described as follows:
\begin{equation}
    \label{eqn:rgcn}
    h_{i}^{l+1} = W_o h_i^{l} + \frac{1}{|\mathcal{N}_{s}(i)|}(\sum_{s\in S^{*}}\sum_{j\in\mathcal{N}_s(i)}W_n h_j^{l})
\end{equation}
Where $h_{i}^{l}$ is the vector representation of atom $a_i$ and layer $l$. $W_o$ and $W_n$ are two learnable weight matrices, and $\mathcal{N}_s(i)$ represent the set of nodes adjacent to $a_i$ through edge type $s$.

The layer-wise operation of our version of relational graph attention and how the attention scores are computed are as follows. The term $\alpha_{ij}^{s}$ is the amount of attention that node $a_i$ would pay to its neighbor node $a_j$ (through edge type $s$).
\begin{equation}
    \label{eqn:rgat}
    h_{i}^{l+1} = \alpha_{ii}W h_i^{l} + (\sum_{s\in S^{*}}\sum_{j\in\mathcal{N}(i)} \alpha_{ij} W h_j^{l})
\end{equation}

\begin{equation}
\label{eqn:alpha}
\alpha_{(ij)}^{s} = \frac{\exp(A(W h_i \vert\vert W h_j))}{\sum_{k\in \mathcal{N}_{s}(i)}^{}\exp(A(W h_i \vert\vert W h_k))}
\end{equation}
\vspace{1mm}

To train a reaction-level predictive model, we apply $L \geq 3$ layers of relational graph attention/convolution to the \textit{rxn-hypergraph}. Then final representation of the reaction would be $X = f(x^{r^L}, x^{p^L})$ where $f$ is a selected summarizing function to combine the two sides of the reaction. We use subtraction ($f(x, y) = x - y$) and concatenation ($f(x, y) = x || y$) as the summarizing function. Taking $X$ as the final latent representation of our reaction, we apply a classification or regression head (typically a feed forward MLP) to perform reaction-level classification or regression.  
In the Experiments section, we evaluate the viability of our proposed method by performing several reaction-level prediction tasks. 

\subsection{Interpretability}
Applying any form of graph neural networks on the \textit{rxn-hypergraph} would provide an interpretable framework using the standard attribution methods such as Integrated gradient \citep{sundararajan2017axiomatic} and Class Activation Map (CAM) \citep{zhou2016learning}. However, the structure of the \textit{rxn-hypergraph} is capable of providing a more in-depth interpretation. The attention weights obtained from applying GAT can be used as a measure of importance for propagating information relevant to the predictive task. Given that, we define three types of interpretations for the predictions of a GAT model trained using \textit{rxn-hypeergraph}. (1) The \textit{atom-rxn} interpretation: the multiplication of the attention scores ($\alpha$) of the edges in path from an atom to the corresponding \textit{rxn-hypernode} (e.g. $a_u^i \rightarrow m_i^r \rightarrow x^r$). These scores can be considered as the contribution of each atom to the final representation of the reaction. (2) The \textit{node-node} interpretation: the average of the attention scores of the edges over all the GAT layers used in architecture. These scores can measure the importance of bonds between adjacent atoms in the final representation of the reaction. Also the relative importance of the reacting molecules in the final representation of the reaction can be measured by \textit{node-node} scores of the edges between the \textit{mol-hyernodes} and their corresponding \textit{rxn-hypernode} (e.g. $m_i^r \rightarrow x^r$). This score can indicate the difference between reactants and reagents.
(3) The intermolecular \textit{atom-atom} interpretation: the multiplication of the attention scores of the edges in the path between pair of atoms of different molecules (e.g. $a_u^i \rightarrow m_i^r \rightarrow m_j^r \rightarrow a_v^j$). These scores can be considered as a measure of the pair-wise correlation between atoms. 

\section{Experiments}
To evaluate the viability of the hypergraph representation in various different circumstances, we perform three experiments on different reaction-level predictive tasks using multiple datasets of chemical reactions.

\subsection{Data}
For the first experiment, we use the dataset of chemical reactions from the US patents office (USPTO)\citep{lowe2012extraction} to train a model to classify these reactions into the top 50 highly populated classes of chemical reactions. All of these reactions were classified according to reaction classes presented in \citet{carey2006analysis, roughley2011medicinal} and the RSC's RXNO ontology \citep{kraut2013algorithm} using the NameRxn tool. Similar to the classification experiment presented in \citet{schneider2015development}, we randomly sample 1000 reactions for each of the 50 reaction classes.
Then we split each class into a subclass of 200 random reactions for training
and 800 for testing. This results in 10,000 training reactions and 40,000 test reactions which are uniformly distributed over 50 classes of chemical reactions.

For the second experiments, we use an in-house curated dataset of reaction mechanisms (reaction with a single transition state). This dataset consists of three classes of mechanistic reactions: (1) over 11,000 polar reactions. The reactions wherein a pair of electrons would transfer from an electron donor orbital to an electron acceptor orbital; (2) 2800 of radical reactions. The reactions that involve a radical species; and (3) 2600 pericyclic reactions. The reactions wherein the transition state has a cyclic geometry. We split this dataset into a 80 percent training dataset and 10 percent validation and 10 percent holdout for final testing.

Lastly, we redo the reaction-level experiment described in \citet{fooshee2018deep} as the reaction ranker stage. In this stage, the authors in \citet{fooshee2018deep} used a set of over 11,000 productive polar mechanistic reactions to train a ranker system which ranks a reaction mechanism based on their thermodynamic plausibility. 

\subsection{Reaction Representations}
For each experiment, we compare the performance of different models that are trained using the corresponding reaction representation. These representations which are introduced in the Related Work section are: (1) \textit{reactionFP} representation based on AP, Morgan2, and TT fingerprints; (2) the representations from the transformer models on reaction SMIRKS, \textit{rxnfp} (both pre-trained and trained from scratch); and (3) representations from the relational graph convolution/attention using \textit{rxn-hypergraph}.

\subsection{Training and Hyperparameter Optimization}
We train the graph attention/convolution layers using either cross-entropy or mean squared error, depending on the task. We use the ADAM optimizer \citep{kingma2017adam} and anneal the learning rate with an exponential schedule across the training duration. Training is performed across 4 GPUs for a period of 500 epochs for each of the experiments explained below.

Additionally, we used SHERPA \citep{hertel2020sherpa} to optimize the hyperparameters associated with each predictive task, guided by Bayesian Optimization for each parameter. Specifically, we optimized the number of graph layers $L$, size of the latent representation of nodes $D$, learning rate, learning rate decay, and $L2$ weight regularization term. The final parameters for each experiment are presented in Table \ref{tab:hyperparams}.

\subsection{Classification of USPTO reactions}
The results of this classification experiment are reported in Table \ref{tab:uspto_classification}. Both transformer representation (\textit{rxnfp}) and RGAT on \textit{rxn-hypergraph} achieve the highest accuracies. It is important to mention that this classification scheme is highly dependent on the presence of certain molecules and compounds on the reactant side of the reaction. Such textual dependencies are not representing the underlying chemistry of the reaction which implies that a highly accurate model that uses the text representations might not learn the actual chemistry of the reactions and take advantage of the presence of absence of these textual signatures. 

\begin{table}[t]
\normalsize
\centering
    \caption{Comparing testing accuracy of different representation on a test set of 40,000 chemical reactions from USPTO dataset of chemical reactions.}
    \begin{tabular}{ccl}
    \toprule
    Representation&Network&Accuracy\\
    \hline
        \multirow{3}{*}{\textit{reactionFP}}&AP&0.854\\
        &Morgan2&0.850\\
        &TT&0.852\\
        \cmidrule(lr){1-3}
        \multirow{2}{*}{Transformers}&pre-trained \textit{rxnfp}&0.862\\
        &\textit{rxnfp}&0.925\\
        \cmidrule(lr){1-3}
        \multirow{2}{*}{\textit{rxn-hypergraph}} & RGCN & 0.909  \\
                                                 & RGAT & \bf 0.928 \\
    
    \bottomrule
    
    \end{tabular}

    \label{tab:uspto_classification}
\end{table}

\subsection{Classification of Mechanistic Reactions}
Here we classify the reactions into three classes of polar, radical, and pericyclic reactions. Since the classification scheme is only based on the pair of reacting orbitals (i.e. single transition states), the underlying chemistry might not be complicated for models to learn. Nevertheless, the graph attentional models on \textit{rxn-hypergraph} are outperforming other models and representations. 

\begin{figure}[tb!]
    \centering
    \includegraphics[width=0.9\columnwidth]{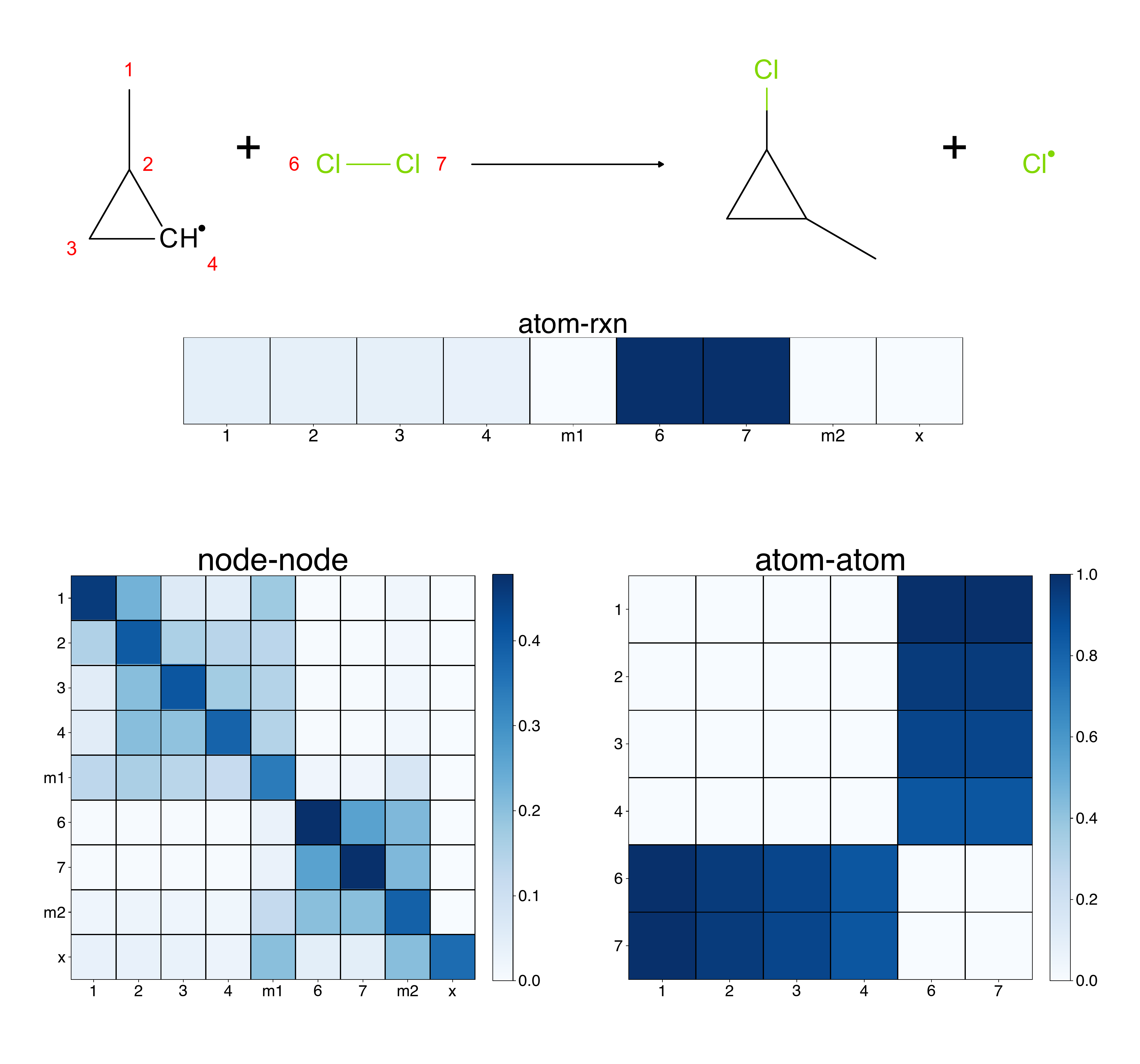}
    \caption{Interpretability plots for a radical reaction for the task of classifying the mechanistic reactions.}
    \label{fig:interpret_radical}
\end{figure}

\begin{figure}[tb!]
    \centering
    \includegraphics[width=0.9\columnwidth]{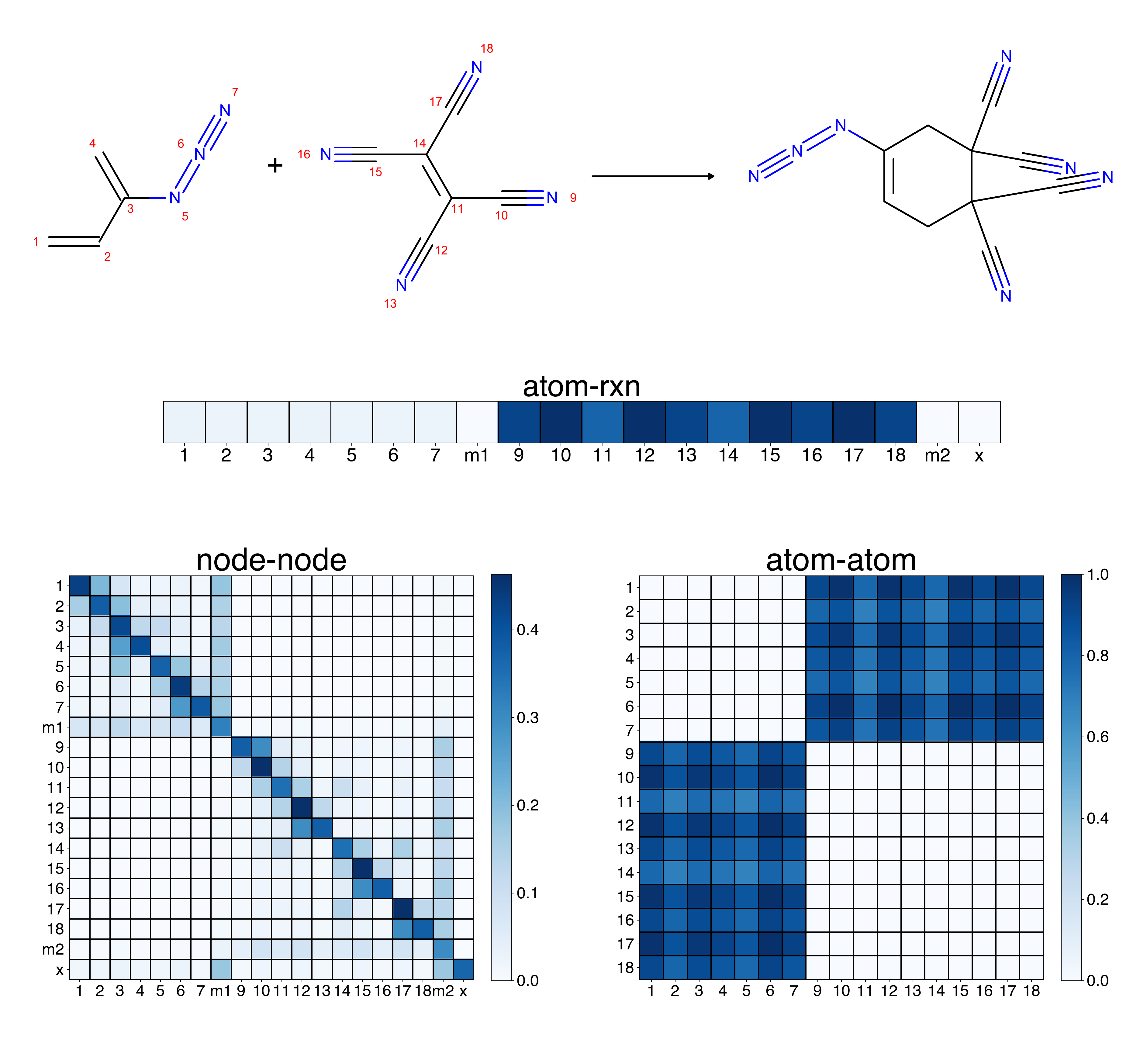}
    \caption{Interpretability plots for a pericyclic reaction for the task of classifying the mechanistic reactions.}
    \label{fig:interpret_peri}
\end{figure}

\begin{figure}[tb!]
    \centering
    \includegraphics[width=0.9\columnwidth]{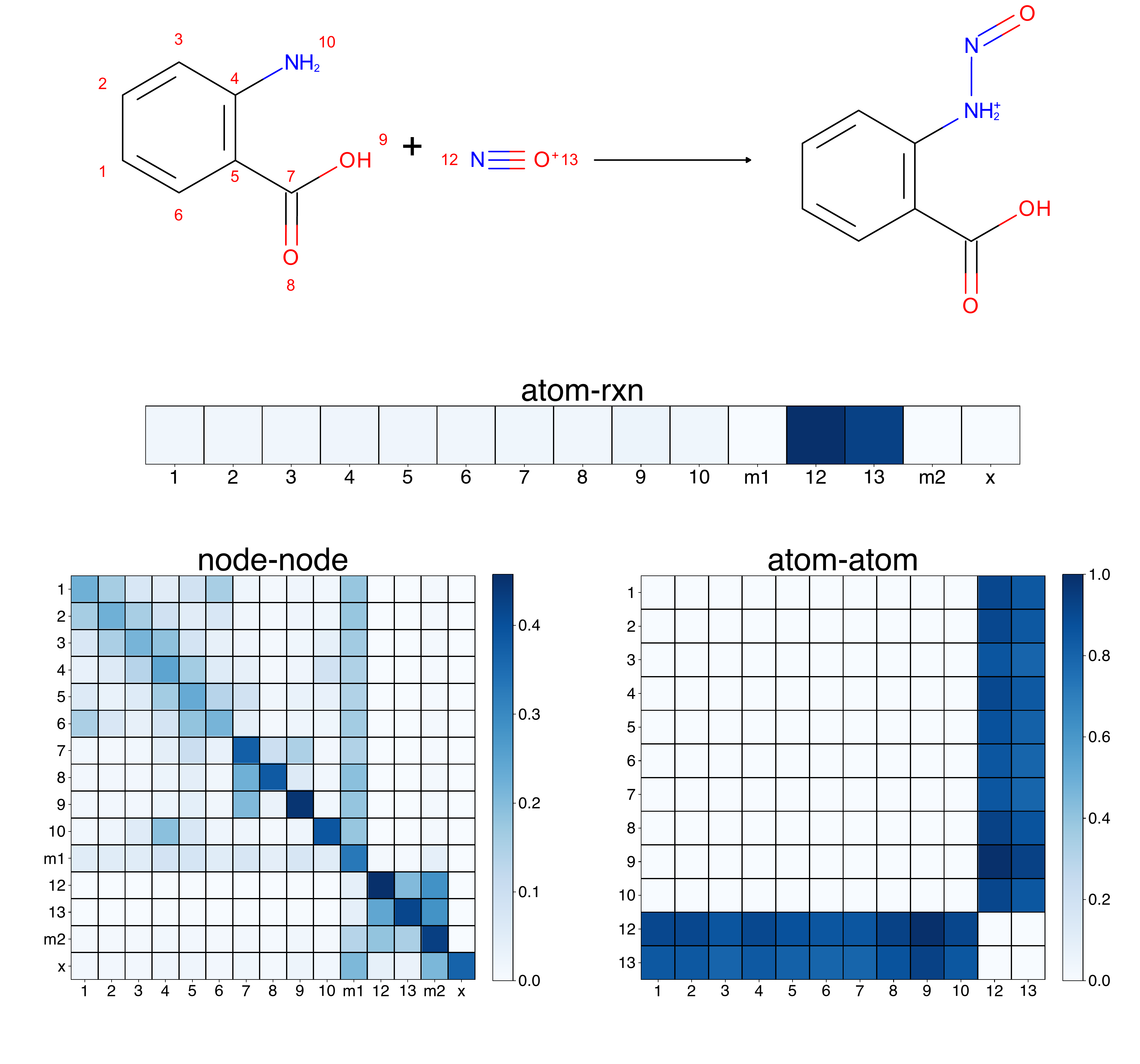}
    \caption{Interpretability plots for a polar reaction for the task of classifying the mechanistic reactions.}
    \label{fig:interpret_polar}
\end{figure}

For this particular experiment, we also show the interpretations of the final predictions using the three metrics described in the Interpretability section. Figures \ref{fig:interpret_radical}, \ref{fig:interpret_peri}, and \ref{fig:interpret_polar} are illustrating the interpretation results. In each figure the reaction with labeled atoms is depicted at the top. The \textit{atom-rxn} scores are shown in the middle while the \textit{node-node} and intermolecular \textit{atom-atom} scores are shown at the bottom right and bottom left of the figures.


\subsection{Plausibility Ranking of Polar Mechanistic Reactions}
This experiment was first introduced in \citet{fooshee2018deep}, where they rank a set of possible reactions from the interactions between one set of reactant molecules. We precisely follow the procedure of ranking described in \citet{fooshee2018deep}, we use Siamese network \citep{bromley1993signature, chicco2021siamese} to train ranker models for \textit{reactionFP} and transformer models. In each branch of the Siamese network, we used the hyperparameters used in the previous section.
\subsubsection{Ranking Network Architecture}
We learn pairwise rankings between different mechanisms using the DirectRanker architecture introduced in \citet{directranker}. We first compute the learned latent representations of a pair of reactions. Afterward, we use the DirectRanker method of subtracting the two latent representations before feeding them through a bias-free fully-connected layer and a sign-preserving anti-symmetric nonlinearity which produces a scalar value for each pair of events. We train this network using mean squared error to predict which of the two input reactions is more plausible than the other, targets of $1$ or $-1$. 

After training this pairwise ranker, we still need a method for determining a final ranked order on a list of plausible mechanisms. To accomplish this, we use the ranked-pair voting method \citep{ranked_pairs}. This algorithm allows us to convert a pair-wise ranking matrix between all viable reaction mechanisms to an ordered list by preference. Ranked pairs construct a directed acyclic graph based on the sorted pair-wise score between different elements. The acyclic property is maintained by ignoring any pairs which would introduce a cycle. After all of the pairs are exhausted, we produce the final ordering by following a topological ordering on the nodes of the graph. The resulting ordering is guaranteed to obey certain criterion which is useful for this task such as independence to irrelevant alternatives, which means that extra implausible reactions which the network is unsure of will not spoil the top rankings.

\begin{table}[t]
\normalsize
\centering
    \caption{Comparing testing accuracy of different representation on a test set of 5,455 chemical reactions from an in house dataset of mechanistic reactions.}
    \begin{tabular}{ccl}
    \toprule
    Representation&Network&Accuracy\\
    \hline
        \multirow{3}{*}{\textit{reactionFP}}&AP&0.915\\
        &Morgan2&0.915\\
        &TT&0.913\\
        \cmidrule(lr){1-3}
        \multirow{2}{*}{Transformers}&pre-trained \textit{rxnfp}&0.955\\
        &\textit{rxnfp}&0.974\\
        \cmidrule(lr){1-3}
        \multirow{2}{*}{\textit{rxn-hypergraph}}& RGCN &  0.988 \\
                                                & RGAT & \bf 0.990  \\
    
    \bottomrule
    \end{tabular}
 \label{tab:mech_classification}
\end{table}

\begin{table*}[!]
\normalsize
\centering
    \caption{The top1, 2, 5, and 10 prediction accuracy of plausibility ranking models with different representations of chemical reaction. All the metrics are computed for a test set of 200 real world mechanistic reactions.}
    \begin{tabular}{ccllll}
    \toprule
    Representation&&top1&top2&top5&top10\\
    \hline
        \multirow{3}{*}{\textit{reactionFP}}&AP&58.01&67.05&77.60&84.33\\
        &Morgan2&59.03&69.14&78.24&85.01\\
        &TT&58.41&68.22&77.31&84.14\\
        \cmidrule(lr){1-6}
        \multirow{2}{*}{Transformers}&pre-trained \textit{rxnfp}&81.31&84.19&89.60&92.33\\
        &\textit{rxnfp}& \bf 89.14&93.22&96.09&98.55\\
        \cmidrule(lr){1-6}
        \multirow{2}{*}{\textit{rxn-hypergraph}}&RGCN& 82.67 & 92.57 & 97.03 & 99.01\\
        & RGAT & 84.23 & \bf 96.06 & \bf 98.57 & \bf 99.28\\
    
    \bottomrule
    \end{tabular}
 \label{tab:ranking}
\end{table*}
Out of three experiment, ranking reaction based in thermodynamic plausibility requires a deeper understanding of the underlying chemistry. In this task, the correlation between the textual signatures and plausibility of a reaction is minimal. This potentially explains the results presented in Table \ref{tab:ranking} where the RGAT model on \textit{rxn-hypergraph} is outperforming other models especially the models based on the text representations with a considerable margin.

\begin{table*}[!]
    \centering
    \begin{tabular}{l|ccc}
        \toprule
        \textbf{Task} & \textbf{USPTO Classification} & \textbf{Mechanism Classification} & \textbf{Polar Plausibility} \\
        \hline
        Number of Graph Attention Layers &  10 & 10 & 5\\
        Latent Representation Dimension &  128 & 128 & 64  \\
        Learning Rate & 4.11 $\times 10^{-4}$ & 4.11 $\times 10^{-4}$ & 1.14 $\times 10^{-2}$\\
        Learning Rate Decay & 0.999995 & 0.999995 & 0.99986\\
        $L2$ Weight Regularization & 9.75 $\times 10^{-5}$ & 9.75 $\times 10^{-5}$ & 7.28 $\times 10^{-5}$ \\
        \bottomrule
    \end{tabular}
    \caption{Table of hyperparameters selected through Bayesian Optimization using SHERPA \citep{hertel2020sherpa}.}
    \label{tab:hyperparams}
\end{table*}

\section{Discussion and Conclusion}
We proposed \textit{rxn-hypergraph} representation of a chemical reaction which is suitable for training graph neural networks for reaction-level predictive task. The key idea behind forming the \textit{rxn-hypergraph} is to construct efficient message passing routes which provides a platform for (1) updating atom representation based on the atom and molecules if the other reacting molecules, and (2) a global pooling mechanism. \textit{Rxn-hypergraph} is designed to be a universal and permutation invariant representation that adapt to any downstream predictive task. There is no manual and hand-crafted pre-processing stages involved in computing the \textit{rxn-hypergraph} and it provides different levels of interpretability. There are two potential aspects of this work which are left to the future work: (1) several other demanding and useful reaction-level predictive tasks such as yield prediction, and reaction rate constant prediction that can be benefited by \textit{rxn-hypergraph}, and (2) more complicated and expressive attention mechanisms such as multiplicative attentions and transformers can be applied to \textit{rxn-hypergraph} which might result in more powerful reaction representations.

\bibliographystyle{unsrtnat}
\bibliography{main}

\end{document}